\documentclass[nohyperref]{article}

\usepackage{microtype}
\usepackage{graphicx}
\usepackage{subfigure}
\usepackage{placeins}
\usepackage{booktabs} 

\usepackage{hyperref}

\usepackage[accepted]{icml2023}

\usepackage{amsmath}
\usepackage{amssymb}
\usepackage{mathtools}
\usepackage{amsthm}

\providecommand{\myparagraph}[1]{\noindent\textbf{#1. }}
\providecommand{\sectionvspace}{\vspace{-.1cm}}

\usepackage[capitalize,noabbrev]{cleveref}

\theoremstyle{plain}

\theoremstyle{definition}

\theoremstyle{remark}

\usepackage[textsize=tiny]{todonotes}

\icmltitlerunning{Distilling VLMs into Embodied Agents}

\begin{document}

\twocolumn[
\icmltitle{Distilling Internet-Scale Vision-Language Models into Embodied Agents}

\icmlsetsymbol{equal}{*}

\begin{icmlauthorlist}
\icmlauthor{Theodore Sumers}{princeton,equal}
\icmlauthor{Kenneth Marino}{deepmind}
\icmlauthor{Arun Ahuja}{deepmind}
\icmlauthor{Rob Fergus}{deepmind}
\icmlauthor{Ishita Dasgupta}{deepmind}

\end{icmlauthorlist}

\icmlaffiliation{princeton}{Department of Computer Science, Princeton University, Princeton, New Jersey}
\icmlaffiliation{deepmind}{DeepMind, New York City, United States}

\icmlcorrespondingauthor{Theodore Sumers}{sumers@princeton.edu}
\icmlkeywords{Machine Learning, ICML}

\vskip 0.3in
]



\printAffiliationsAndNotice{\icmlEqualContribution} 

\begin{abstract}
Instruction-following agents must ground language into their observation and action spaces. 
Yet learning to ground language is challenging, typically requiring domain-specific engineering or large quantities of human interaction data. 
To address this challenge, we propose using pretrained vision-language models (VLMs) to supervise embodied agents. 
We combine ideas from model distillation and hindsight experience replay (HER), using a VLM to retroactively generate language describing the agent's behavior. 
Simple prompting allows us to control the supervision signal, teaching an agent to interact with novel objects based on their names (e.g., planes) or their features (e.g., colors) in a 3D rendered environment.
Fewshot prompting lets us teach abstract category membership, including pre-existing categories (food vs toys) and ad-hoc ones (arbitrary preferences over objects).
Our work outlines a new and effective way to use internet-scale VLMs, repurposing the generic language grounding acquired by such models to teach task-relevant groundings to embodied agents.

\end{abstract}
\sectionvspace
\section{Introduction}
\sectionvspace
\label{sec:introduction}
Embodied agents capable of understanding and fulfilling natural language instructions are a longstanding goal for artificial intelligence~\cite{winograd1972understanding}.  
Such agents must \emph{ground} language~\cite{harnad1990symbol, mooney2008learning} by correctly associating words with corresponding referents in their environment.
But grounding language is both philosophically~\cite{Quine1960} and practically challenging: methods to learn such groundings remain an active area of research~\cite{tellex2020robots}. 

\begin{figure}[t]
\begin{center}
\centerline{\includegraphics[width=\columnwidth]{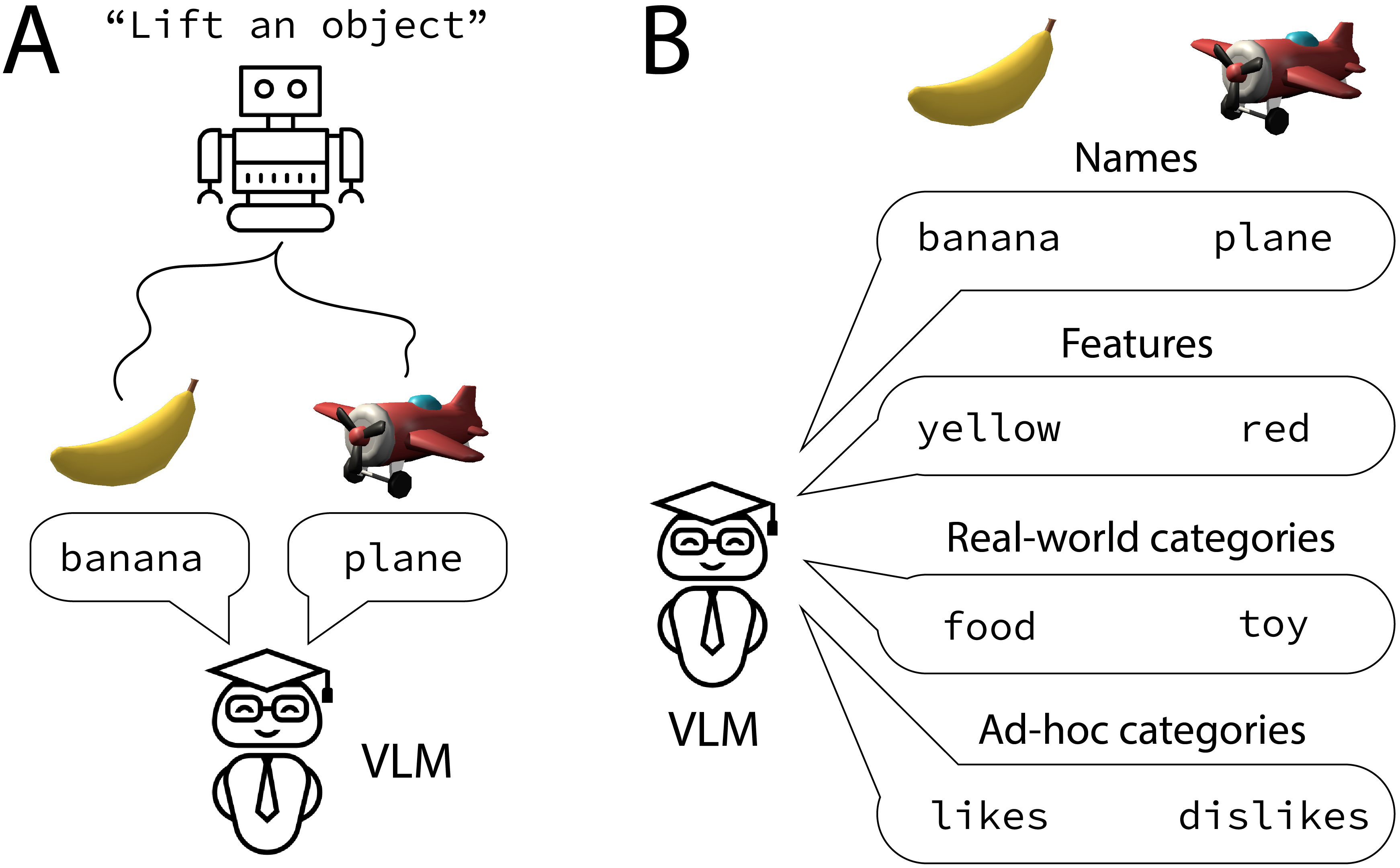}}
\vspace{-.2cm}
\caption{Overview of our approach. \textbf{A}: We use a generative VLM to re-label agent trajectories for hindsight experience replay. \textbf{B}:~Varying the prompt allows us to relabel trajectories along multiple dimensions to teach the embodied agent different tasks.}
\label{fig-intro}
\end{center}
\vspace{-.75cm}
\end{figure}

Embodied language grounding is particularly difficult because training data are scarce. 
Passive learning from internet data has driven a series of revolutions in natural language processing~\cite{mikolov2013linguistic, bert18, Brown2020LanguageMA}, but embodied agents must map language into their own idiosyncratic observation and action spaces.
 
Training data for embodied agents thus typically consist of trajectories paired with linguistic instructions~\citep[][\emph{inter alia}]{tellex2011understanding, abramson2020imitating} or descriptions~\cite{nguyen2021interactive, sharma2022skill, zhong2022improving}. 
Providing an agent with aligned behavior and language allows it to learn a mapping between the two. 
Unfortunately, such data are expensive (typically requiring human annotation), under-specified (a trajectory can correspond to several different instructions or descriptions), and agent-specific (trajectory representations are tightly coupled with the agent architecture).

We propose using large-scale vision-language models (VLMs) to address some of these challenges. Unlike embodied agents, VLMs can be trained on massive internet data~\citep{Radford2021LearningTV, Ramesh2022HierarchicalTI, alayrac2022flamingo, openai2023gpt4, liu2023prismer}. Here, we use the Flamingo VLM~\cite{alayrac2022flamingo} to annotate trajectories, leveraging its internet-derived language grounding to generate training data for the embodied agent (Fig.~\ref{fig-intro}).

Our approach distills the VLM's domain-general language grounding into domain-specific embodied agents, thus ameliorating data cost and scarcity. We use prompting and few-shot learning to control which aspects of a trajectory to relabel, mitigating label under-specification and allowing new task definitions on the fly. 
Finally, unlike approaches which integrate encoders such as CLIP~\cite{Radford2021LearningTV} directly into the embodied agent, our method operates purely as data augmentation. This allows us to remain agnostic to changes in the agent architecture and affords substantial interpretability, making interventions and diagnosis easier.

After reviewing related work (Sec.~\ref{sec:related-work}), we make the following contributions:
\begin{itemize}
\vspace{-.3cm}
\item A novel method using a generative VLM to supervise training of language-conditioned agents (Sec.~\ref{sec:method}).
\vspace{-.25cm}
\item Experiments using our method to flexibly teach new language groundings, including object names (Sec.~\ref{section_exp1}), attributes (Sec.~\ref{section_exp2}), category membership (Sec.~\ref{section_exp3}) and even ad-hoc user preferences (Sec.~\ref{section_exp4}).
\vspace{-.25cm}
\item An analysis of this imperfect supervision signal, including transferable insight into how different types of noise affect downstream task performance (Sec.~\ref{section_label_analysis}).
\vspace{-.3cm}
\end{itemize}
Taken together, our work demonstrates that generic language grounding acquired from internet-scale pretraining can be controlled and distilled into embodied agents, allowing us to teach task-specific language groundings without the burden of extensive human supervision. 

\section{Related Work}
\sectionvspace
\label{sec:related-work}
Our work studies language grounding within the classic instruction following setting, where an embodied agent is given a natural language instruction and must generate a trajectory satisfying it~\cite{winograd1972understanding}.\footnote{Outside instruction following, a related body of work grounds descriptive language from documents~\cite{branavan2012learning, zhong2019rtfm, zhong2021silg} or interactions~\cite{narasimhan2018grounding, sumers2021learning, lin2022inferring} to learn general policies.} 

\sectionvspace
\subsection{Learning from Human Interactions}
\sectionvspace
Typical approaches use a human-generated dataset of aligned language and trajectories to learn a mapping between them~\citep[][\emph{inter alia}]{kollar2010toward, tellex2011understanding, chen2011learning, artzi2013weakly, mei2016listen, anderson2018vision, Blukis2019LearningTM, lynch2020grounding, abramson2020imitating, shridhar2020alfred, fried2018speaker}. 
Language-conditioned reinforcement learning (RL) learns this grounding from trial and error, either assuming an environment-generated reward signal \cite{misra2017mapping, chaplot2018gated, yu2018interactive}, or learning a reward function from aligned language-trajectory data~\cite{bahdanau2018learning}.

While these works developed agents capable of fulfilling natural language instructions within circumscribed domains, they are limited by training on static environments and scarce, expensive datasets. The resulting learned language groundings are often tightly coupled to the agent's observations and actions, 
making them inflexible to new objects or concepts. 
Researchers have explored numerous methods to mitigate this, including data augmentation~\cite{Blukis2020FewshotOG, chen2022learning}, dual-coding memory~\cite{hill2021grounded}, auxiliary language generation objectives~\cite{yan2022intraagent, bigazzi2021exploreexplain}, or interactive supervision~\cite{kulick2013active, mohan2014learning, she2014teaching, thomason2017opportunistic, co2018guiding, chai2018language, nguyen2021interactive}. Recent approaches instead leverage internet-scale language models to achieve generalization.

\sectionvspace
\subsection{Leveraging Pretrained Models}
\sectionvspace
Several lines of work use language-only knowledge from pretrained models. For example, word embeddings can be used to generalize representation of linguistic instructions~\cite{Chen2020AskYH, li2019robust}, while language models can be used to break complex instructions into simpler ones~\cite{ahn2022can, Huang2022LanguageMA, huang2022inner, dasgupta2022collaborating, singh2022progprompt}. Because these approaches use pretrained weights over language only, they can help the agent generalize paraphrases or combinations of existing concepts, but cannot themselves provide groundings for novel visual concepts.

Unlike these language-only models, vision-language models (VLMs) acquire multi-modal grounding from passive internet-scale learning. Contrastive VLMs \citep[e.g. CLIP, ][]{Radford2021LearningTV} can be used to drive exploration~\cite{tam2022semantic}, construct a semantic representation from visual imagery~\cite{chen2022open, singh2022progprompt}, or directly as the agent's vision and text encoders, thus inheriting this grounding~\cite{majumdarzson2022, shahlm2022, Shridhar2021CLIPortWA, Bucker2022LaTTeLT}.
However, such pretrained vision encoders are often trained for object recognition and may not encode task-relevant information \citep[e.g., spatial positions, ][]{Shridhar2021CLIPortWA}. Further, since this information is encoded as hidden vectors, it is nontrivial to determine which representation will be optimal for a particular downstream task~\cite{hsu2022what} or avoid biases arising from their training data~\cite{bommasani2021opportunities}. Our approach---using generative VLMs to produce linguistic annotations---is inherently interpretable and allows us to use simple prompts to focus the VLM on task-relevant aspects of the visual scene.

Finally, other approaches have trained new VLMs on internet-scale data to serve as general~\cite{nair2022r3m} or domain-specific~\cite{Fan2022MineDojoBO, guhur2021airbert, hao2020towards} visual representations. In contrast, our approach uses an off-the-shelf pretrained VLM to generate task-relevant annotations.

\sectionvspace
\subsection{Hindsight Experience Replay}
\label{section_HER_background}
\sectionvspace
Our approach builds on Hindsight Experience Replay~\citep[HER,][]{andrychowicz2017hindsight} in a language-conditioned setting~\cite{chan2019actrce}. HER's key insight is that early in training, rewards are sparse because agents rarely (if ever) achieve the specified goal. HER densifies rewards by converting these failed trajectories into successful ones by retroactively ``relabeling'' them, assigning a goal achieved by the agent's behavior. 

The central challenge of applying HER to language-conditioned agents is implementing a relabeling function that maps from states to natural language instructions. First, the mapping is not 1:1, as multiple instructions may be compatible with the same state (for example, ``Lift a banana'' and ``Lift something yellow'', Fig.~\ref{fig-intro}). Second, the mapping is not known: there is no clear way of converting an agent's state to a linguistic instruction fulfilled by that state. 

Prior work has used a reward signal to train a domain-specific relabeling model~\cite{cideron2020higher}, a mix of image- and human-relabeling~\cite{lynch2020grounding}, or thousands of human-relabeled trajectories to fine-tune a contrastive VLM which can then be used to select the best instruction from a candidate set~\cite{xiao2022robotic}. We instead use a pretrained generative VLM as the relabeling function, omitting the need for environmental rewards or domain-specific human labelled data. Our approach also offers free-form language generation (rather than selecting from limited options) and on-the-fly task specifications via prompting (allowing us to relabel specific aspects of the trajectory, such that each can have multiple language labels depending on the task specification; Fig.~\ref{fig-intro}B). This allows better use of trajectories. 

\begin{figure*}[t!]
\begin{center}
\centerline{\includegraphics[width=17cm]{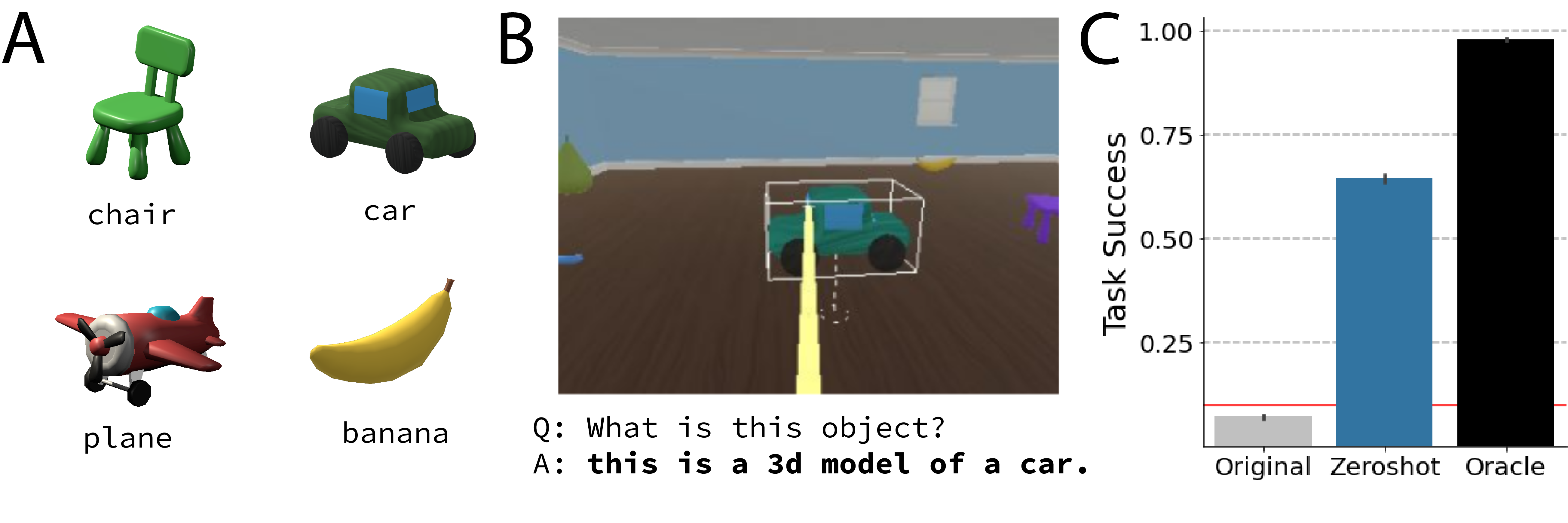}}
\caption{Using a pretrained VLM to teach an agent novel object names (Sec.~\ref{section_exp1}). \textbf{A}: Four of the ten objects. The agent has never seen these words before, and the VLM is not given any information about what might be present. \textbf{B}: We use an open-ended prompt to relabel trajectories, then use the VLM outputs (bold) to retrain the agent. \textbf{C}: Results from the original agent (``Original''), after retraining with VLM labels (``Zeroshot''), and an upper bound from retraining on ground truth labels from the environment (``Oracle''). Here and throughout, the red line shows baseline performance (lifting a random object) and error bars show 95\% CIs.}
\label{fig-exp1-summary}
\end{center}
\vspace{-.75cm}
\end{figure*}

\sectionvspace
\section{Method}
\label{sec:method}
\sectionvspace
 Our approach uses a pretrained generative VLM as the relabeling function for HER (Sec.~\ref{section_HER_background}). We first formally specify HER. HER assumes that goals $g\in\cal{G}$ correspond to predicates $f_g : \cal{S} \to \{\text{0,1}\}$, so that goals are completed by reaching a set of states. During training, an agent is provided with a goal $g$ and executes a trajectory $\xi = s_0, ... s_T$. Because the goal is rarely accomplished, most trajectories receive no reward: $\forall_{s \in \xi} f_g(s) = \text{0}$. HER solves this by \emph{relabeling} the trajectory, assigning a new goal $g'$ which \emph{was} accomplished by the last state in the sequence: $f_{g'}(s_T) = \text{1}$.\footnote{Alternative HER formulations relabel randomly sampled states. In our setting, the last state is generally the most informative, but applications to other domains such as vision-language navigation may benefit from such strategies.} The relabeling function $m$ takes a state as input and returns a goal fulfilled by that state $m : \cal{S} \to \cal{G} \text{ s.t. } \forall_{\text{s} \in \cal{S}} \text{f}_{\text{m(s)}}(\text{s}) = \text{1}$.
 
 In our work, we use a VLM as the relabeling function. We relax the assumption that the environment is fully observable, and consider \emph{observation} sequences $o_1,..., o_T$ (the agent's visual inputs) generated by an observation function $O: \cal{S} \to$ $\Omega$. The VLM serves as a proxy relabeling function $\tilde{m}$, which now takes an observation and returns a goal (a natural language string) satisfied by the state underlying that observation: $\tilde{m} : \Omega \to \cal{G} \text{ s.t. } \forall_{\text{s} \in \cal{S}} \text{f}_{\tilde{\text{m}}\text{(O(s))}} (\text{s}) = \text{1}$. Intuitively, this means that we retroactively generate language that describes the agent's actual behavior (Fig.~\ref{fig-intro}A). 

This approach is conceptually straightforward and fully compatible with any procedure for training language-conditioned agents, as the original instructions can be directly replaced by the generated ones. Indeed, while HER was developed for reinforcement learning, we use the relabeled trajectories to train with imitation learning instead.

Relative to prior work, our approach makes few assumptions and is highly data efficient. Unlike~\citet{cideron2020higher} we do not use a reward signal from the environment; and unlike~\citet{lynch2020grounding, xiao2022robotic} we do not crowdsource relabeling. However, in settings with a reward signal or preexisting datasets, it would be possible to incorporate such information into our method (e.g., using an annotated dataset to fine-tune the generative VLM). 

Finally, while we use a generative VLM, object classification or detection models \cite{minderer2022owlvit, kuo2022f, kirillov2023segment} may be used instead. Such models have two important structural limitations: they require \emph{a priori} specification of possible goals (unlike our zero-shot experiments in Sections~\ref{section_exp1}-\ref{section_exp3}) and cannot be used for ad-hoc categories (Section~\ref{section_exp4}). Empirically, we found that the classification / detection paradigm was less effective than our generative approach: substituting the OWL-ViT detection model \cite{minderer2022owlvit} as a relabeling function yielded substantially worse performance (Appendix~\ref{appdx-sec-owlvit}). 

\sectionvspace
\section{Experimental Setup}
\label{section_experimental_setup}
\sectionvspace
In this section, we describe our experimental framework employing a VLM to relabel trajectories for HER. Following the classical HER formulation~\cite{andrychowicz2017hindsight} we use a simple task structure which can be re-labeled on the basis of the final observation only. We return to extensions requiring multiple observations in the discussion.

\sectionvspace
\subsection{The Playhouse Environment}
\sectionvspace
We use the Playhouse environment from~\citet{abramson2020imitating}, a Unity-based environment with a continuous action space. Each episode takes place in a procedurally-generated home with multiple rooms and a wide range of everyday domestic objects (Fig.~\ref{fig-exp1-summary}A). This 3D environment is challenging for RL agents and the VLM, as the agent's egocentric perspective often yields unusual or close-up perspectives on objects (Figs.~\ref{fig-exp1-example-labels},~\ref{fig-exp2-example-labels},~\ref{fig-exp3-example-labels}).

\sectionvspace
\subsection{The ``Lift'' Task}
\sectionvspace
To isolate the effects of relabeling we use the same task structure across our experiments. We chose the ``Lift'' task as it gives a direct measure of our method's effectiveness: we can evaluate the agent's language grounding on an object-by-object basis.

At the start of each episode, agents are placed in a room with 5-10 objects and instructed to lift a target object. To demonstrate the flexibility of our method, we vary the task specification: using object names (``Lift a plane''; Sec~\ref{section_exp1}), attributes (``Lift a red object''; Sec~\ref{section_exp2}), categories (``Lift a toy''; Sec~\ref{section_exp3}), or preferences (``Lift something John Doe likes''; Sec~\ref{section_exp4}). Episodes end when the agent lifts an object, or after 120 seconds. 

\sectionvspace
\subsection{Flamingo VLM}
\sectionvspace
We use Flamingo~\cite{alayrac2022flamingo}, a state-of-the-art language-generative VLM. Flamingo accepts interleaved images and text as input, and produces text output. This allows us to experiment with both ``Zeroshot'' prompts (containing only the image to be relabeled and a text prompt) and ``Fewshot'' prompts (including up to 32 in-context image-text examples). We use the 80B parameter model described by~\citet{alayrac2022flamingo} with greedy sampling.


\sectionvspace
\subsection{HER Implementation}
\sectionvspace
We first need an agent that generates structured behaviors that can be interpretably relabelled. We use human-human data to learn a task-agnostic motor policy: e.g., an agent that knows how to \emph{lift} something, but not what a \emph{plane} is. We refer to this as the ``original'' agent, and train it via behavioral cloning (BC) on the human-human dataset described by~\citet{interactiveagents2021creating}; for details on the dataset, agent architecture, and BC implementation, please refer to that work. To ensure the ``original'' agent lacks task-relevant groundings, we filter out episodes with relevant utterances before performing BC. However, we note that the ``original'' agent could be generated in any way, such as with RL on a different set of tasks; our approach uses it as a starting point.

\begin{figure*}[t!]
\begin{center}
\centerline{\includegraphics[width=17cm]{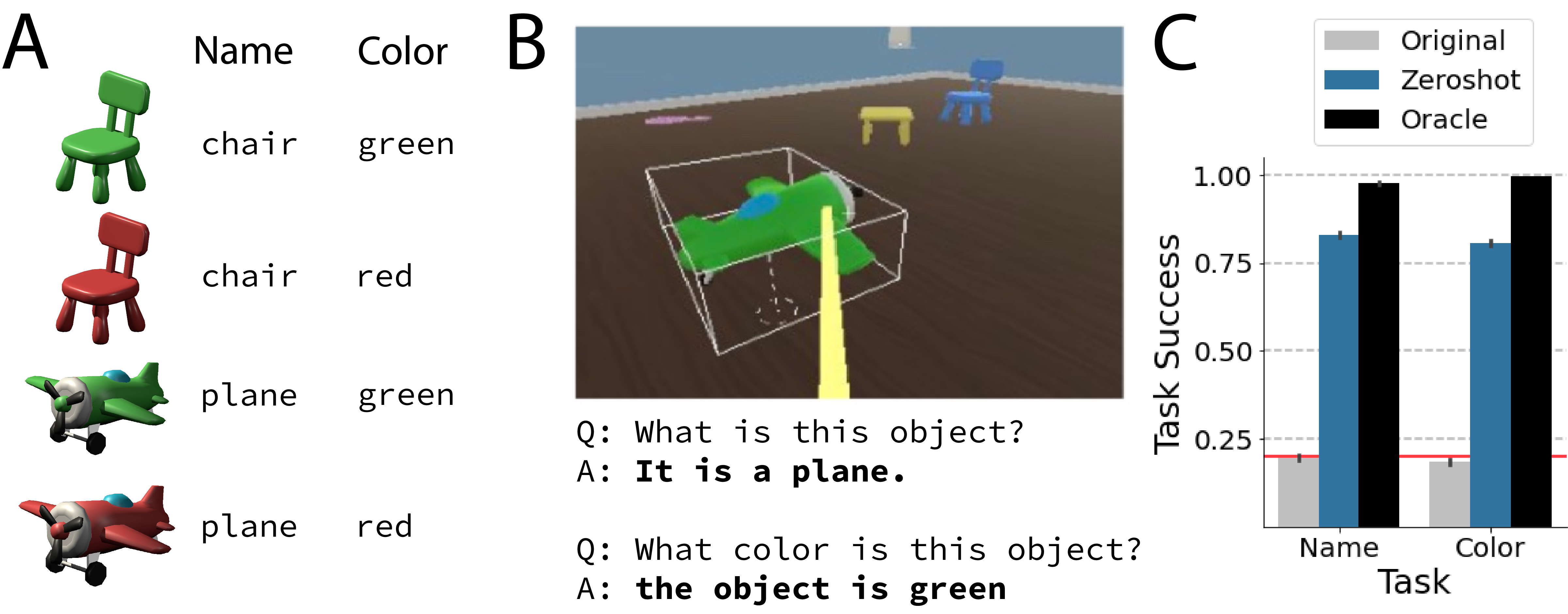}}
\vspace{-.25cm}
\caption{Using a pretrained VLM to flexibly teach object attributes (name or color) from a single set of trajectories (Sec.~\ref{section_exp2}). \textbf{A}: We randomize color mappings so each object can appear red, green, blue, pink, or yellow. \textbf{B}: We again use generic prompts, adding a second color-oriented prompt to obtain color labels. \textbf{C}: Results on the ``Name'' and ``Color'' tasks from the ``Original'' agent and after retraining with VLM labels (``Zeroshot'') or ground truth labels from the environment (``Oracle'').}
\label{fig-exp2-summary}
\end{center}
\vspace{-.75cm}
\end{figure*}

To compare the effects of different relabeling functions, we use a batched HER approach (Fig.~\ref{appdx-pipeline-schematic}). For each experiment, we generate an initial set of approximately 10,000 trajectories with a generic ``Lift an object'' instruction (due to implementation details, the actual number varied from 10,000 to 11,500).
Across all experiments, around 3\% of these initial trajectories timed out as the agent did not lift an object. This was not enough to meaningfully affect results, so for simplicity we discarded them. This is equivalent to assuming the agent can detect a successful grasp, which is reasonable even in robotics \cite{pinto2016grasp}. 

We then use the VLM to relabel the final image in each trajectory. This generates a new annotation describing the agent's actual behavior (Fig.~\ref{fig-intro}A,~\ref{fig-exp1-summary}B). We perform light post-processing on the VLM outputs: Flamingo sometimes generates multiple responses (always separated by a newline), so we truncate the response to the first newline and prepend ``Lift a ''. We then use these VLM-generated strings as the instruction in a second round of BC. Finally, performing BC on the full trajectories is computationally expensive. Preliminary analysis showed that using the full trajectories provides only minor performance gains compared to truncating them to the last 5 seconds; we therefore adopt this truncation throughout.

\sectionvspace
\section{Results}
\label{section_results}
\sectionvspace
To demonstrate the flexibility of VLMs for relabeling, we vary the goal structure of the ``Lift'' task. Our first two experiments focus on visible object attributes, using simple ``Zeroshot'' prompts to teach object names (Sec.~\ref{section_exp1}) and attributes (Sec.~\ref{section_exp2}). Our third and fourth use ``Fewshot'' prompts to teach category membership (Sec.~\ref{section_exp3}) and finally novel user preferences over objects (Sec.~\ref{section_exp4}). We close with an analysis of label noise and task performance (Section~\ref{section_label_analysis}). This highlights an advantage of our approach over contrastive VLMs~\cite{Radford2021LearningTV}: our method produces human-legible language annotations with corresponding confidence scores, allowing us to analyze and filter labels to improve downstream task performance.

\sectionvspace
\subsection{Teaching Object Names}
\label{section_exp1}
\sectionvspace
Our first experiment uses relabeling to teach the agent to lift one of 10 objects: a table, a chair, a book, a basketball, a racket, a plane, a car, a banana, a carrot, and a pear (Fig.~\ref{fig-exp1-summary}A).

\myparagraph{Setup}
We begin by training an ``original'' agent, filtering out any episodes containing one of the 10 target words.\footnote{Due to the tokenizer used in the agent, we filtered out all episodes containing the substring ``ball'' rather than ``basketball.''} We use this agent to generate 10,000 initial trajectories using the generic ``Lift a object'' instruction in a room with all 10 objects. We provide the VLM with the final image in each trajectory and use a simple QA-style zero-shot prompt: \texttt{[IMG\_0] Q: What is this object? A: } (Fig.~\ref{fig-exp1-summary}B). We then re-train the agent on these new VLM labels.

\myparagraph{Results}
We test our retrained agent by again placing it in a room with all 10 objects, but now instructing it to lift a specific one (e.g., ``Lift a car'') and generate 10,000 evaluation trajectories. We first test the ``Original'' agent to ensure it has no knowledge of the objects. We find that it achieves near-chance performance (7.1\% task success), confirming that it does not possess task-relevant language groundings. We then test an ``Oracle'' agent retrained on ground-truth relabeling from the environment itself. This agent performs near ceiling (97.9\% success), confirming there are enough trajectories to teach the task with perfect relabeling. Finally, we find that our VLM-based relabeling method---using only a simple zero-shot prompt and no information about what objects might appear---conveys a significant fraction of the task, resulting in 64.4\% success (Fig.~\ref{fig-exp1-summary}C). The VLM-retrained agent performs well above chance on all ten objects (Fig.~\ref{fig-exp1-per-item-results}).

Intriguingly, we achieve this performance despite substantial noise in the relabeling. VLM-generated strings are relatively low accuracy (only 54.7\% contain the canonical object name used in instructions), and frequently contain extraneous words (Fig.~\ref{fig-exp1-summary}B, \ref{fig-exp1-example-labels}; Table~\ref{appdx-exp1-zeroshot-labels}). We conduct a deeper analysis of label noise and downstream task performance in Sec.~\ref{section_label_analysis}.

\sectionvspace
\subsection{Teaching Object Attributes}
\label{section_exp2}
\sectionvspace
Section~\ref{section_exp1} demonstrated that VLMs can teach basic object names. But how controllable is the VLM relabeling? Can we, for example, teach an agent to recognize an object's \emph{attributes} rather than the objects themselves? We now show that the VLM can be used to relabel a single set of trajectories with their \emph{name} or \emph{color} respectively.

\begin{figure*}[t]
\begin{center}
\centerline{\includegraphics[width=17cm]{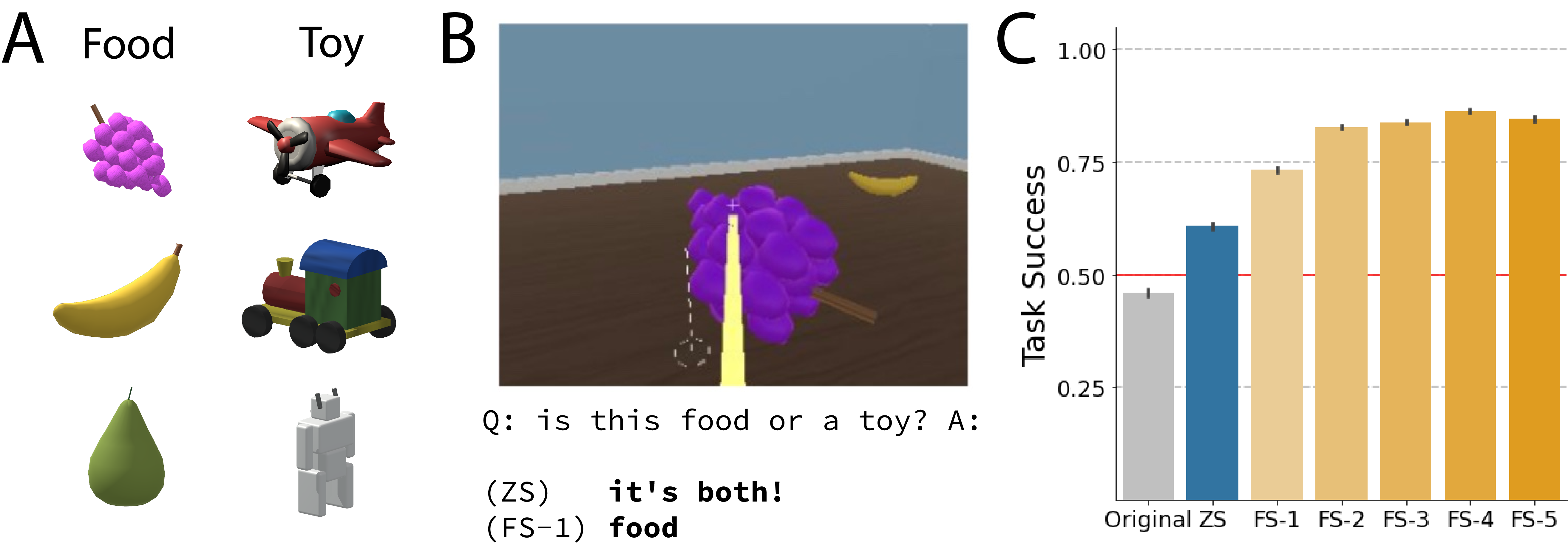}}
\vspace{-.25cm}
\caption{Using a pretrained VLM to teach category structure (Sec.~\ref{section_exp3}). \textbf{A}: We use a set of 10 objects, 5 ``food'' and 5 ``toys.'' \textbf{B}: We experiment with both ``Zeroshot'' (ZS) and ``Fewshot'' (FS) prompting. We vary the number of fewshot examples from each category from one (FS-1 is given 3 examples each of carrots and robots) to all five (FS-5 is given 3 examples each of all 5 objects in each category). \textbf{C}:~Results for the ``Original'' agent and after retraining with different relabelings. Performance increases substantially from Zeroshot to FS-2 then plateaus. This suggests that partial information about the category is sufficient for Flamingo to extrapolate to new food and toys.}
\label{fig-exp3-summary}
\end{center}
\vspace{-.5cm}
\end{figure*}

\myparagraph{Setup}
We use a subset of five objects from Section~\ref{section_exp1} (plane, racket, chair, table, and basketball). Previously, these items' colors were fixed across all episodes. We now render them in different colors (red, green, blue, pink, or yellow). This gives us a total of 25 object-color combinations (green chair, red chair, green plane, etc.; Fig.~\ref{fig-exp2-summary}A).  As before, we train an original agent filtering out episodes containing any of these object name or color terms. We create a level with all five objects, and randomly assign each color to a object in each episode. We use this task to generate a set of 10,000 trajectories with a generic ``Lift an object'' prompt. 

Now, however, we relabel each trajectory twice (Fig.~\ref{fig-exp2-summary}B,~\ref{fig-exp2-example-labels}). One relabeling uses the original prompt from Section~\ref{section_exp1}: \texttt{[IMG\_0] Q: What is this object? A: }. The second introduces a slight variation, adding the word ``color'': \texttt{[IMG\_0] Q: What color is this object? A: }.  We follow the same procedure and re-train two separate agents: one using the labels generated by the original prompt, and one using the ``color'' variation.

\myparagraph{Results}
We again test our agents by placing them in a room with all five objects and instructing them to lift one.  However, we now test two forms of instructions: an object name task (``Lift a \{plane, racket, chair, table, basketball\}'') or an object color task (``Lift a \{red, green, blue, pink, yellow\} object''). We use the same comparisons, checking the original agent and an agent retrained on oracle color and object name labels.

We again find that the original agent performs at chance (19.6\% task success on names and 18.4\% on colors) and the oracle-relabeled agent performs at ceiling (97.6\% on names and 99.5\% on color). Our VLM relabeling achieves 83.0\% and 80.6\% respectively (Fig.~\ref{fig-exp2-summary}C).

\sectionvspace
\subsection{Teaching Real-World Categories}
\label{section_exp3}
\sectionvspace
Sections~\ref{section_exp1} and~\ref{section_exp2} show the VLM can be used to re-label visual object attributes such as shape and color. But many important properties, such as category membership, are not readily visible. We next test whether our method can be used for tasks depending on such properties. 

\myparagraph{Setup}
We use a set of 10 items, five ``food'' (pear, banana, carrot, lemon, and grapes) and five ``toys'' (plane, train, car, robot, dice; Fig.~\ref{fig-exp3-summary}A). We train an original agent, filtering out episodes with references to any of these, as well as ``food'' or ``toy.'' Again, we generate a set of initial trajectories using a ``Lift an object'' prompt in a room with all 10 objects.

We again use a simple prompt to re-label the trajectories: \texttt{[IMG\_0] Q: Is this food or a toy? A: }. However, initial results suggested that these categories caused an interaction effect with the 3D rendered graphics: Flamingo recognized that the ``food'' items were really \emph{toy} food items and frequently labeled them as ``toys'' or ``both'' (Fig.~\ref{fig-exp3-summary}B, Table~\ref{appdx-foodtoy-zeroshot-labels}). This domain shift issue could be obviated by using a different category structure (e.g. food vs furniture), but we instead experimented with few-shot prompting.
We generated three example images of the agent lifting each of the ten items and ran a series of relabelings with incrementally more examples. Our ``Fewshot-1'' prompt gives Flamingo three examples each of one food and one toy (carrots and robots) for a total of 6 in-context examples. ``Fewshot-2'' added lemons and dice (a total of 12 examples); ``Fewshot-3'' added planes and bananas (total of 18); ``Fewshot-4'' added grapes and cars (total of 24), and finally ``Fewshot-5'' added trains and pears for a total of 30. ``Fewshot-5'' therefore saw three examples each of all 10 objects. Few-shot relabeling resulted in short VLM generations with no extraneous words (Fig.~\ref{fig-exp3-summary}B,~\ref{fig-exp3-example-labels}, Table~\ref{appdx-foodtoy-fewshot-labels}).

\begin{figure}[t]
\begin{center}
\vspace{-.1cm}
\centerline{\includegraphics[width=\columnwidth]{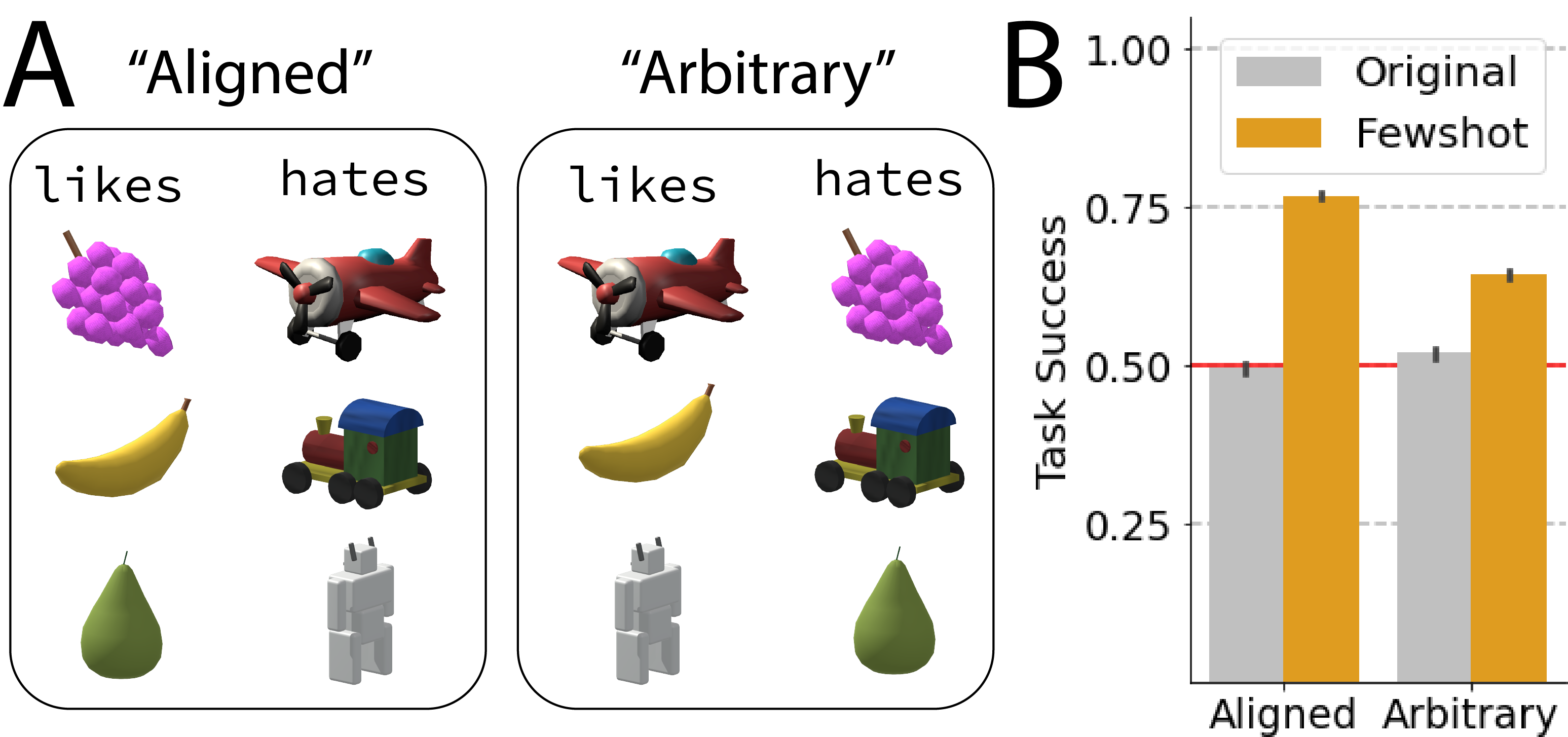}}
\vspace{-.3cm}
\caption{Using a VLM to teach ad-hoc categories (Sec.~\ref{section_exp4}). \textbf{A}:~``Aligned'' preferences follow existing category structure, while ``Arbitrary'' preferences cut across it. \textbf{B}: The VLM is able to teach tasks requiring new category structure from fewshot examples, but alignment with existing structure helps.}
\label{fig-exp4-summary}
\end{center}
\vspace{-.75cm}
\end{figure}

\begin{figure*}[t]
\begin{center}
\centerline{\includegraphics[width=17cm]{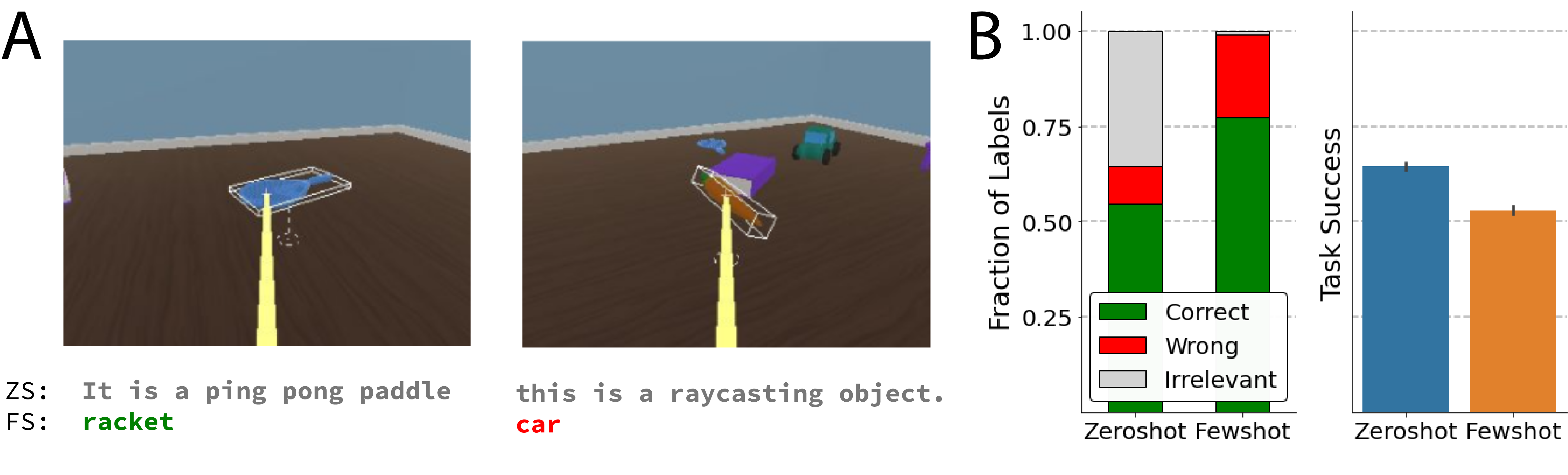}}
\vspace{-.5cm}
\caption{Comparing ``Zeroshot'' (ZS) and ``Fewshot'' (FS) relabeling (Sec.~\ref{section_label_analysis}). \textbf{A}: Zeroshot typically generates text that reflects the image contents, but is often not task-relevant. Fewshot prompting encourages the VLM to generate one of the 10 object labels. However, when the foreground is challenging it often labels background objects instead. \textbf{B}: Label quality and task performance. Zeroshot yields many irrelevant labels; Fewshot instead produces more ``Correct'' and ``Wrong'' labels. Agents trained on Fewshot labels perform worse.}
\label{fig-FS-ZS-intro}
\end{center}
\vspace{-.5cm}
\end{figure*}

As before, we re-train agents with each of the resulting label sets. We evaluate them by generating 10,000 rollouts each for ``Lift a food'' and ``Lift a toy'' instructions.

\myparagraph{Results}
As expected, our original agent performs near chance (46.0\% task success). Zeroshot relabeling lifts performance above chance to 60.8\%. Adding in-context examples provides another substantial boost: ``Fewshot-1'' lifts performance to 73.3\%, and ``Fewshot-2'' to 82.8\% (Fig.~\ref{fig-exp3-summary}C). Adding additional examples has only a marginal effect, with performance plateauing around 85\%. Notably, even in the ``Fewshot-2'' condition, Flamingo is readily able to generalize category structure to the remaining three items in each category. This suggests that the few-shot examples help Flamingo adapt to the 3D rendered visuals, while Flamingo inherits information about category membership from its pretraining experience.

\sectionvspace
\subsection{Teaching Ad-hoc Categories}
\label{section_exp4}
\sectionvspace
Our first three experiments show that simple prompting and a handful of few-shot examples allow our VLM to recognize and teach canonical properties such as names, colors, and category membership. In general, however, we cannot expect tasks to conform to existing canonical categories. Evidence from psychology suggests that \emph{ad-hoc categories} \cite{barsalou1983ad} play a crucial role in human cognition, allowing us to create new conceptual groupings on the fly. Such context-dependent categories are often based on usecases (``things to take on a camping trip'') or affordances (``things that can be used as firewood in an emergency''). Can our method be used to teach such flexible category structures?

Our final experiment tests the VLM's ability to re-label based on new and arbitrary category structure: here, instantiated as a user's preferences over a set of objects. Such dynamic relabeling would allow individuals to provide personalized task specifications. For example, a user could provide a list of items to bring on a camping trip, or express preferences (or allergies) over food items. The agent could then re-label its previous experience with such items and re-train its policy to learn the new groundings, allowing it to map user-level requests into its action space.

\myparagraph{Setup}
We aim to test Flamingo's ability to learn (and then teach) new categories via in-context examples. We re-use the 10 objects from Sec.~\ref{section_exp3} but re-formulate the task in terms of user preferences. We introduce two sets of preferences: ``Aligned'' preferences, which respect existing category structure (John Doe likes food and dislikes toys), and ``Arbitrary'' preferences, which cut across it (John Doe likes robots, planes, carrots, lemons, and bananas; and dislikes cars, dice, trains, grapes, and pears; Fig.~\ref{fig-exp4-summary}A). 

We use the 10,000 rollouts generated in Section~\ref{section_exp3} but re-label them with new prompts: \texttt{[IMG\_0] Q: Would John Doe like this? A:}. We include preambles \texttt{John Doe likes food.} for ``Aligned'' and \texttt{John Doe likes robots, planes, carrots, lemons, and bananas.} for ``Arbitrary''. We use fewshot examples for all 10 items (equivalent to ``Fewshot-5'' from Sec.~\ref{section_exp3}) with ``yes'' or ``no'' responses according to the category structure being used. To produce task-appropriate labels, we transform Flamingo's response by mapping ``yes'' to ``an object John Doe likes'' and ``no'' to ``an object John Doe hates.''

We re-train agents using these labels, and again evaluate by averaging 10,000 rollouts on the two tasks: ``Lift something John Doe likes'', and ``Lift something John Doe hates.''

\myparagraph{Results}
Our original agent performs near chance (50\%) for both category structures (Fig.~\ref{fig-exp4-summary}B). We find that Flamingo relabeling improves performance for both category structures, with ``Aligned'' (76.7\%) outperforming ``Arbitrary'' (64.2\%). These results demonstrate that Flamingo is aided by, but not solely dependent on, real-world category structure. 

It is helpful to compare these results with the ``Fewshot-5'' results from Section~\ref{section_exp3}. These experiments each use the same few-shot examples to relabel the same trajectories, but vary the nature of the category structure.
Flamingo does very well with \emph{explicit} real-world category structure (``Food-Toy'' in Sec.~\ref{section_exp3}; task performance 84.7\%); next best with \emph{implicit} real-world category structure (``Aligned'' in Sec.~\ref{section_exp4}; 76.7\%); and modestly with \emph{no} real-world category structure (``Arbitrary'' in Sec.~\ref{section_exp4}; 64.2\%). 

Finally, we tested the VLM's ability to generalize categories to new instances by permuting the toys' colors. We found that for both preference structures, the VLM successfully relabeled recolored objects with equivalent accuracy (Appendix~\ref{appdx-sec-recoloring-generalization}).

\sectionvspace
\subsection{Analyzing the Relabeling Function}
\label{section_label_analysis}
\sectionvspace
In this final section, we use the setup from Section~\ref{section_exp1} to conduct an analysis of Flamingo as a relabeling function.

\sectionvspace
\subsubsection{Zeroshot vs Fewshot Flamingo}
\sectionvspace

We conduct a ``Fewshot'' relabeling version of ``Zeroshot'' experiments in Section~\ref{section_exp1}, with 32 examples (3-4 examples each of the 10 items in the task) included in context.
We find that ``Fewshot'' relabeling increases accuracy, but---surprisingly---\emph{decreases} downstream task performance. Concretely, 77.2\% of ``Fewshot'' labels contain the canonical object name (compared to 54.7\% of ``Zeroshot''), yet the ``Fewshot''-retrained agent achieves only 52.9\% task success (compared to 64.4\% for the ``Zeroshot'' agent; Fig.~\ref{fig-FS-ZS-intro}). 

A closer look at the labels reveals an important difference. ``Zeroshot'' labels that are not correct are often \emph{irrelevant} (i.e. they do not contain any of the 10 task-relevant object names). In contrast, incorrect ``Fewshot'' labels are almost always \emph{wrong} (i.e., they contain a different task-relevant object name). Few-shot prompting encourages Flamingo's generation towards task relevant labels, causing it to ``guess'' a label when uncertain (Fig.~\ref{fig-FS-ZS-intro}B,~\ref{fig-exp1-example-labels},~\ref{appdx-exp1-fs-zs-unigram-statistics}). While ``Zeroshot'' Flamingo is noisier, these irrelevant labels have little effect on downstream performance. In contrast, ``Fewshot'' generates wrong task-relevant labels that actively interfere with grounding. 

Because many ``Zeroshot'' relabelings are irrelevant, ``Fewshot'' is higher \emph{accuracy} ($\frac{\text{\# of correct labels}}{\text{\# of trajectories}}$) but lower \emph{precision} ($\frac{\text{\# of correct labels}}{\text{\# of task-relevant labels}}$). 
Therefore ``Zeroshot'' Flamingo produces more \textit{reliable} data: more examples relabeled ``Lift a car'' will actually reflect the appropriate behavior. 

\begin{figure*}[t]
\begin{center}
\centerline{\includegraphics[width=17cm]{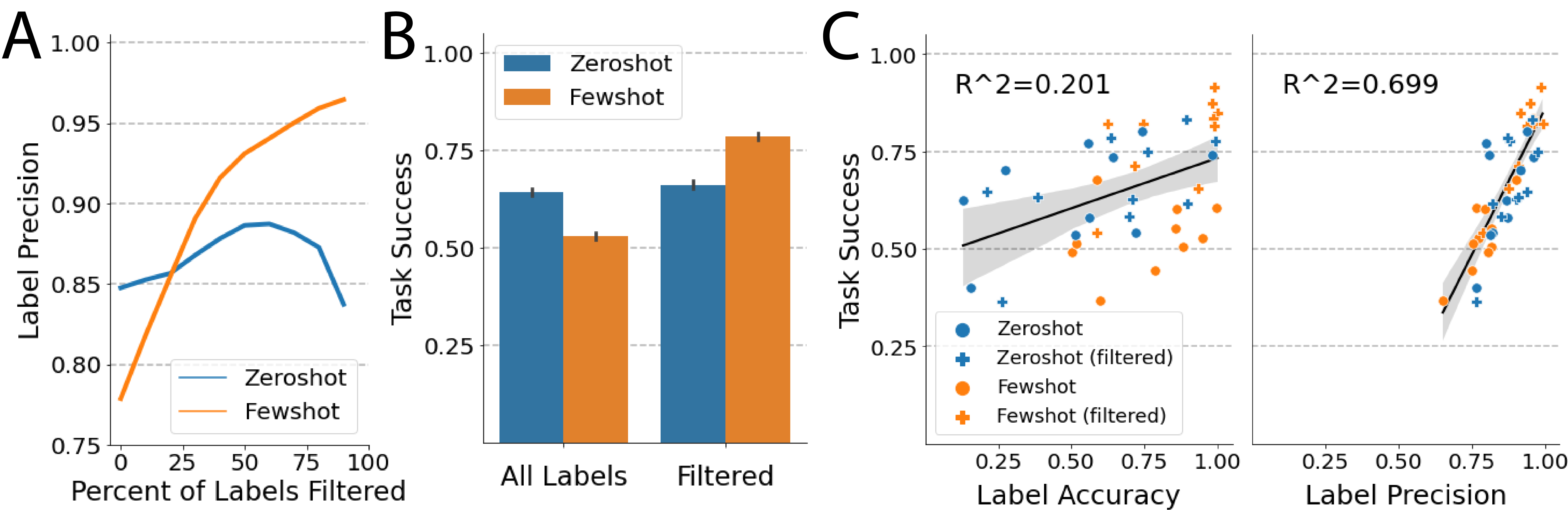}}
\vspace{-.25cm}
\caption{Analysis of label noise and downstream performance (Sec.~\ref{section_label_analysis}). \textbf{A}: Filtering out low-confidence labels dramatically increases ``Fewshot'' label precision, but only marginally improves ``Zeroshot.'' \textbf{B}: Filtering out the least confident 50\% of labels and re-training agents results in a substantial performance gain for ``Fewshot'' but not ``Zeroshot.'' \textbf{C}: Analysis of label characteristics against downstream task performance on a per-object basis suggests that relabeling precision is more important than accuracy.}
\label{fig-FS-ZS-filtering}
\end{center}
\vspace{-.5cm}
\end{figure*}

While initial results suggest that ``Fewshot'' is actually a worse relabeling function, we find that ``Fewshot'' provides an important benefit: it helps calibrate Flamingo's confidence in its relabeling (Fig.~\ref{fig-label-analysis-histograms}). We experiment with filtering out low-confidence labels by progressively dropping the lowest decile. We find that this filtering dramatically increases ``Fewshot'' precision, but only slightly increases ``Zeroshot'' precision (Fig.~\ref{fig-FS-ZS-filtering}A). We filter out the least-confident 50\% of the labels and retrain agents on the remaining trajectories. This substantially improves ``Fewshot'' performance (from 52.9\% to 78.6\%) while only marginally improving ``Zeroshot'' (64.4\% to 66.1\%; Fig.~\ref{fig-FS-ZS-filtering}B).

\sectionvspace
\subsubsection{Label Properties and Task Performance}
\sectionvspace
These results suggest that relabeling \emph{precision} may be more important than \emph{accuracy} for HER. We quantify this by looking at task success resulting from different label sets.

Formally, we are interested in whether label accuracy or label precision is a better predictor of downstream task performance. We have 10 tasks (each of the individual objects used in Section~\ref{section_exp1}, i.e. ``Lift a car'', ``Lift a plane''), and four sets of labels for each (Zeroshot, Zeroshot filtered, Fewshot, and Fewshot filtered). We use a mixed-effects linear regression to predict task success for each of these 40 data points, with fixed effects of label precision and recall, and random effects for each of the 10 tasks. We find that both accuracy and precision are significant, but the effect size of precision is nearly ten times that of accuracy (accuracy: $\beta = .16, t(35.88)=3.416, p<.01$; precision: $\beta = 1.44, t(35.76)=11.5, p<1e-10$; see Table~\ref{appdx-regression-results}). Fig~\ref{fig-FS-ZS-filtering}C plots regression lines for accuracy and precision respectively. 
This result yields the valuable general insight that  
relabeling precision is more important than relabeling accuracy for downstream task performance with HER.

\sectionvspace
\section{Discussion}
\sectionvspace
In this work, we used a VLM pretrained on internet data to teach an embodied agent language groundings. We used prompting and fewshot learning to guide the VLM's text generation, focusing it on specific dimensions of the visual stream. This allows us to flexibly distill task-relevant subsets of the VLM's language grounding into the embodied agent.

We note several limitations to our work. First, we focused on English instructions; future work could experiment with pretrained translation models to develop multilingual grounded agents. Second, we demonstrated our method within the classic HER formulation~\cite{andrychowicz2017hindsight} and thus used a task structure that permits relabeling of the final observation only. We additionally used the dataset from~\citet{interactiveagents2021creating} to learn a low-level motor policy before using our method to teach task semantics. Future work could extend our method beyond traditional HER, using a VLM to annotate observation pairs or full videos. Such extensions would facilitate teaching temporally-extended tasks, making our method suitable for training motor policies in addition to task semantics (e.g., teaching both \emph{how} to lift an object and \emph{which} object to lift). Finally, the Flamingo model~\cite{alayrac2022flamingo} used in this work is not publicly available. However, the recently released GPT-4 \cite{openai2023gpt4} and open-source Prismer models provide comparable capabilities \cite{liu2023prismer}. Object classification or detection models \cite{minderer2022owlvit} may also be used to replicate a subset of our method's functionality.

Future work may experiment with other ways to leverage pretrained generative VLMs for embodied agents. Following our offline supervision approach, VLMs could provide a reward signal during training on the basis of vision and text alignment. Alternatively, VLMs could be used directly in the agent. While large model size\footnote{Our VLM~\cite{alayrac2022flamingo} contains 80B parameters, while our agent~\cite{interactiveagents2021creating} contains 57M.} might mitigate their use for low-level motor policies, they could be used to produce language observations for input to other models~\cite{zeng2022socratic, huang2022inner, dasgupta2022collaborating}. Our method may also be straightforwardly applied to vision-language navigation, another popular testbed for embodied agents \cite{anderson2018vision}. Overall, the confluence of more naturalistic environments~\cite{shridhar2020alfred, savva2019habitat, li2021igibson} with strong and flexible pretrained VLMs~\cite{alayrac2022flamingo, liu2023prismer, openai2023gpt4} makes these models an appealing source of domain-general language groundings. We hope that our method spurs further research leveraging their strengths for embodied agents.

\section*{Acknowledgements}
We thank Christine Kaeser-Chen, Nathaniel Wong, Fede Carnevale, Alexandre Fréchette, Felix Hill, and the DeepMind NYC team for their advice and assistance. 
\bibliography{main}
\bibliographystyle{icml2023}

\newpage
\appendix
\onecolumn

\renewcommand{\thefigure}{S\arabic{figure}}
\renewcommand{\thetable}{S\arabic{table}}
\setcounter{table}{0}
\setcounter{figure}{0}

\section{Appendix}

\begin{figure}[h]
\begin{center}
\centerline{\includegraphics[width=14cm]{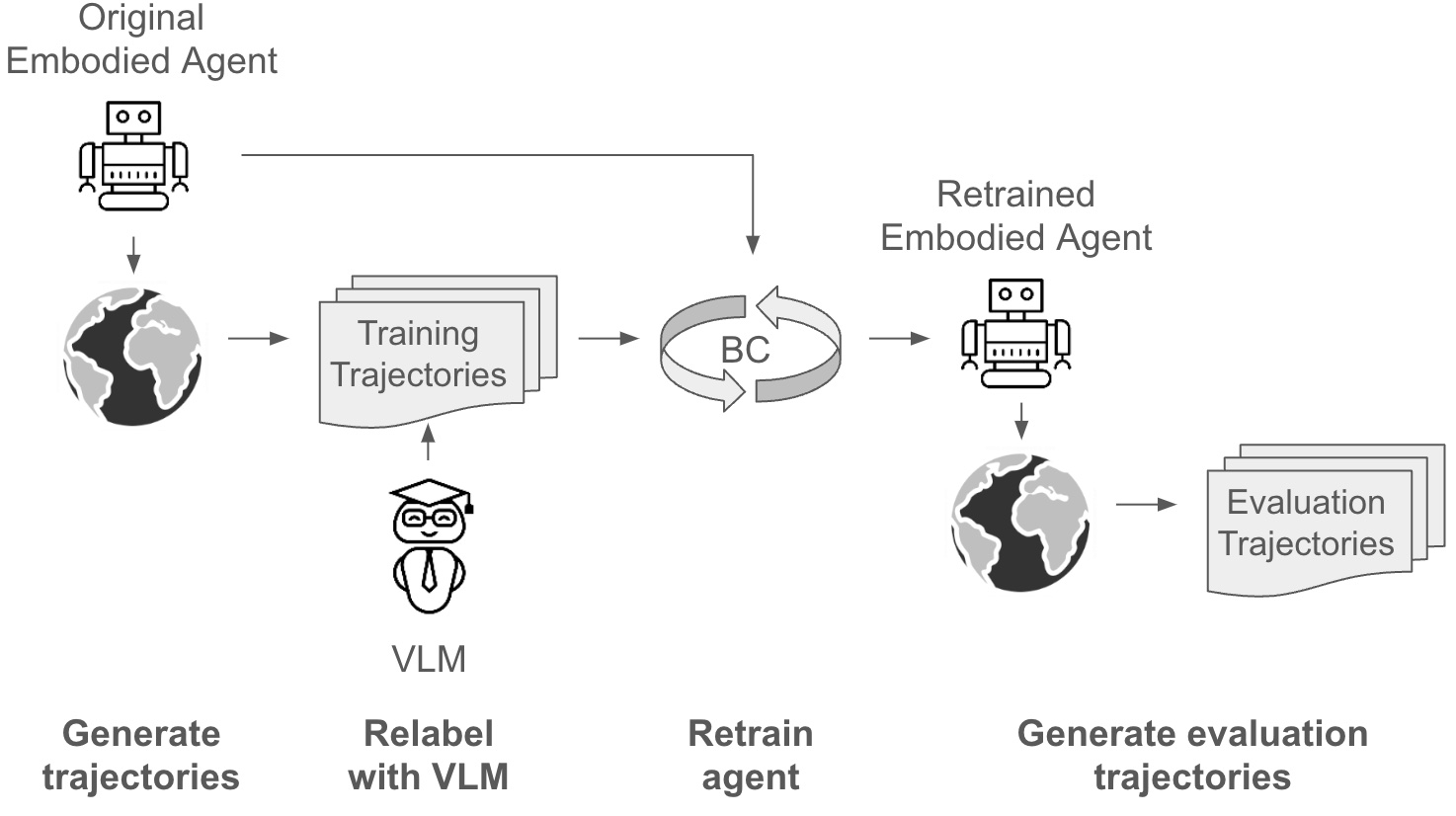}}
\caption{Schematic of our batched HER implementation (Section~\ref{section_experimental_setup}). We generate a batch of trajectories, relabel those trajectories with a VLM, and then re-train the agent with behavioral cloning. We then rollout the retrained agent on target tasks to evaluate performance.}
\label{appdx-pipeline-schematic}
\end{center}
\end{figure}

\begin{figure}[h]
\begin{center}
\centerline{\includegraphics[width=\columnwidth]{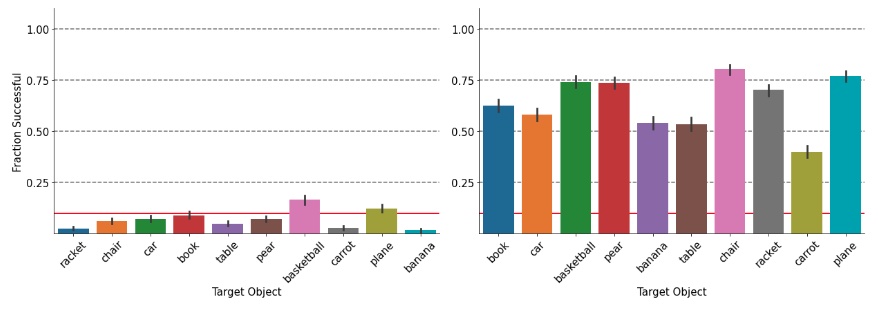}}
\caption{Original and retrained task success per object (Section~\ref{section_exp1}).}
\label{fig-exp1-per-item-results}
\end{center}
\end{figure}

\begin{figure*}[h]
\begin{center}
\centerline{\includegraphics[width=17cm]{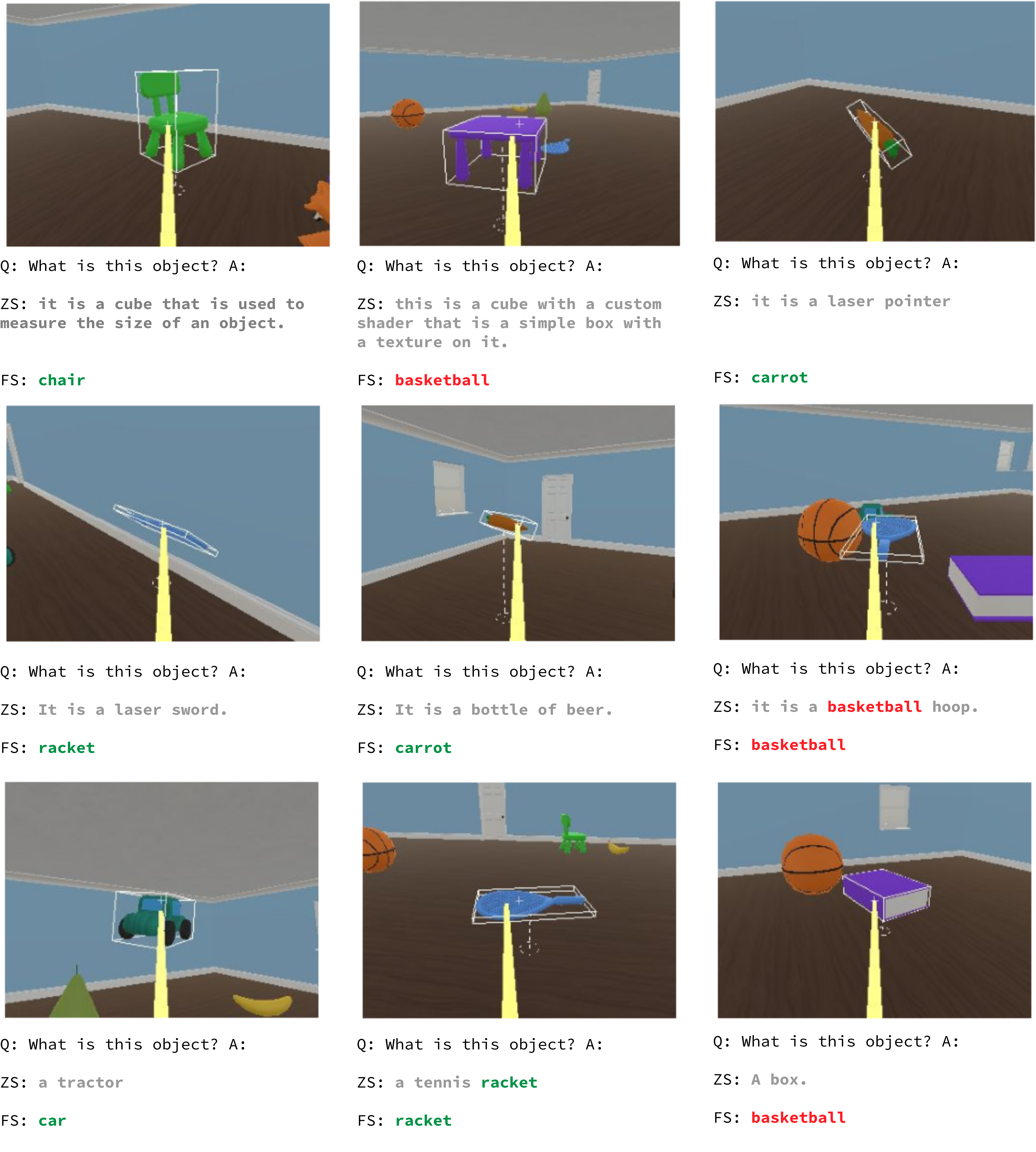}}
\caption{Additional ``Zeroshot'' (ZS, Section~\ref{section_exp1}) and ``Fewshot'' (FS, Section~\ref{section_label_analysis}) relabeling examples. Regular text is the prompt and bold text is the VLM generation. ``Correct'' relabelings are green, ``Wrong'' relabelings are red, and ``Irrelevant'' text is gray. ``Zeroshot'' relabeling generally results in reasonable text, but is distracted by unusual visual features (such as the cube that appears around objects, top left; or the yellow ray indicating the agent is grasping something; top right).}
\label{fig-exp1-example-labels}
\end{center}
\end{figure*}

\begin{figure*}[t]
\begin{center}
\centerline{\includegraphics[width=15cm]{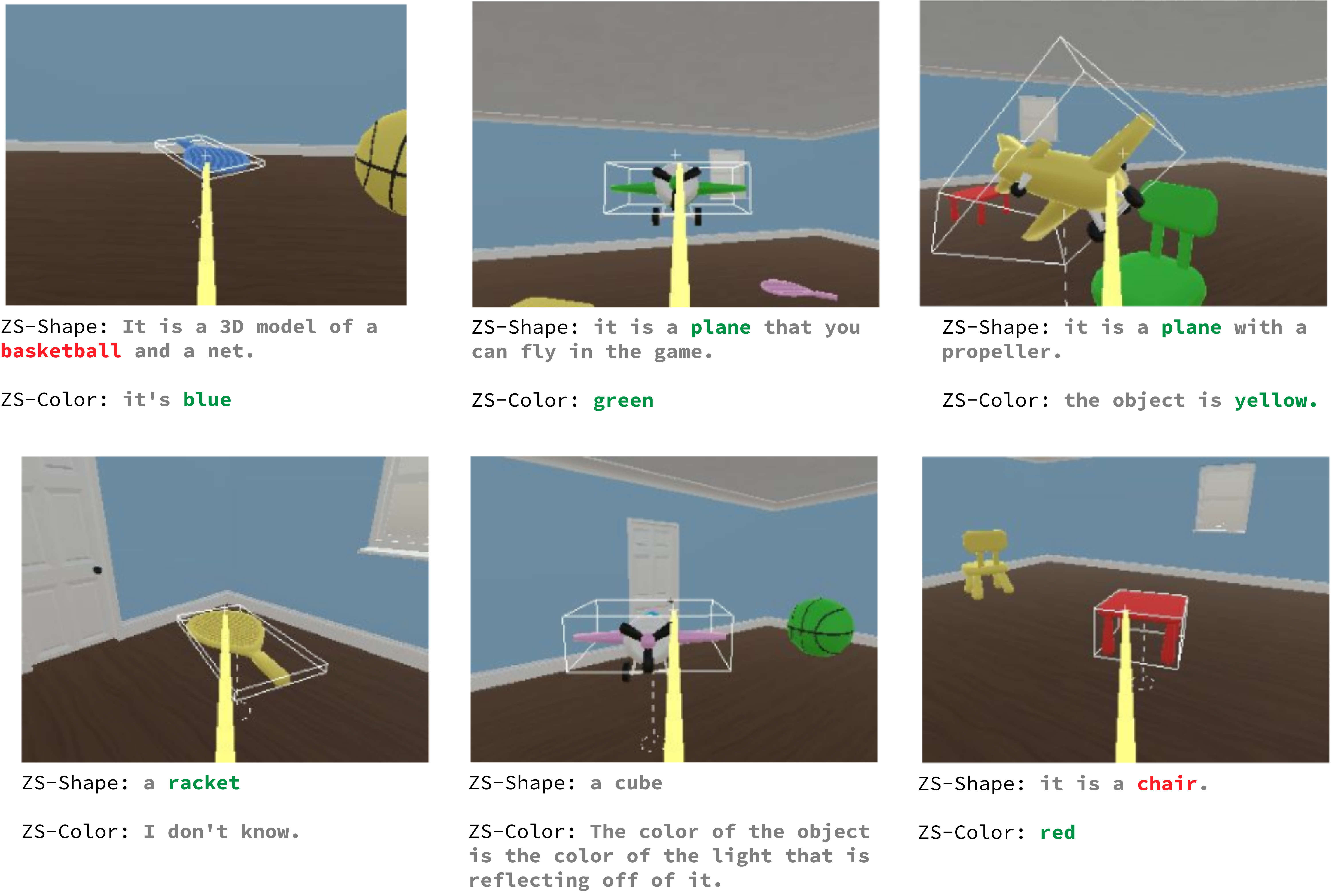}}
\caption{Additional examples for Section~\ref{section_exp2}. VLM relabelings using zeroshot ``names'' (top) and ``color'' (bottom) prompts. Regular text is part of the prompt and bold text is the VLM generation. ``Correct'' relabelings are green, ``Wrong'' relabelings are red, and ``Irrelevant'' text is gray. We conduct parallel VLM-based relabelings of the same initial trajectories in order to teach the agent to recognize \emph{objects} or \emph{colors} respectively.}
\label{fig-exp2-example-labels}
\end{center}
\end{figure*}

\begin{figure*}[t]
\begin{center}
\centerline{\includegraphics[width=16cm]{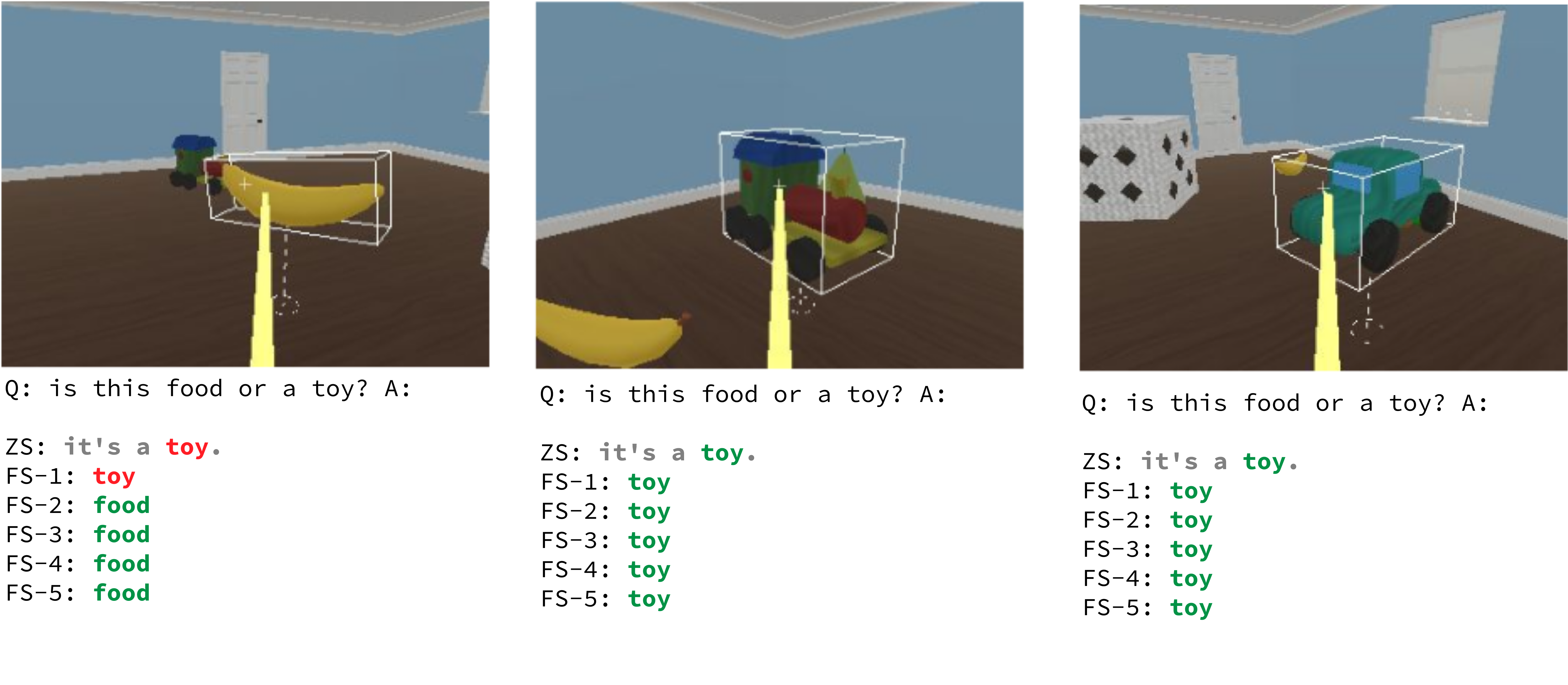}}
\caption{Additional examples for Section~\ref{section_exp3}. The basic prompt template was the same for all; few-shot Flamingo provided in-context examples of food or toy items. Ironically, zero-shot Flamingo claimed that most food items were either ``toys'' or some variant of ``both'', likely due to their 3D rendered appearances (Table~\ref{appdx-foodtoy-zeroshot-labels}). Providing in-context examples readily overcame this issue (Table~\ref{appdx-foodtoy-fewshot-labels}).}
\label{fig-exp3-example-labels}
\end{center}
\end{figure*}

\begin{figure}[t]
\begin{center}
\centerline{\includegraphics[width=10cm]{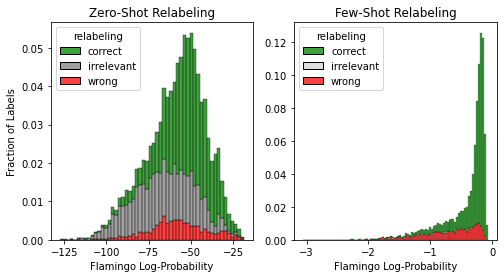}}
\caption{Stacked histograms showing label quality as a function of Flamingo's confidence. For both, wrong or irrelevant labels tend to be low confidence. However, ``Fewshot'' is better calibrated: virtually all correct labels are high-confidence.}
\label{fig-label-analysis-histograms}
\end{center}
\end{figure}

\begin{figure}[t]
\begin{center}
\centerline{\includegraphics[width=10cm]{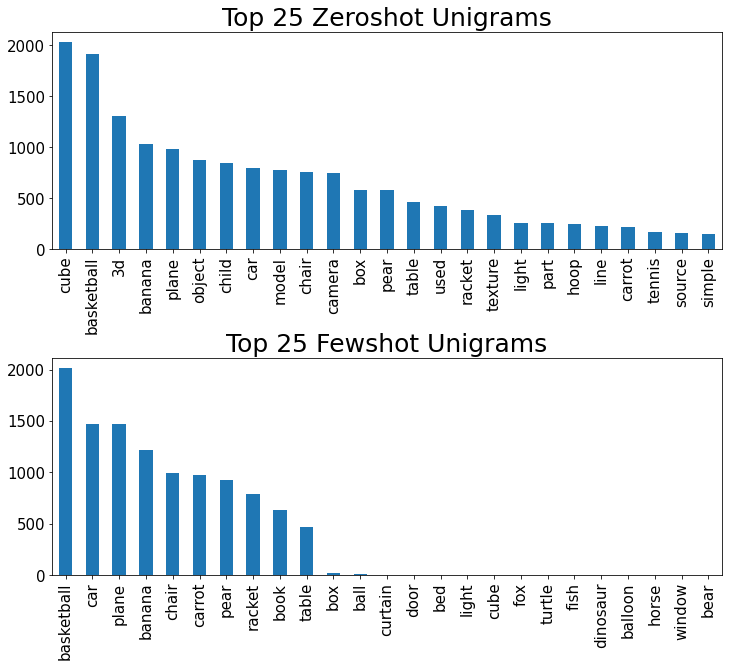}}
\caption{Unigram frequency (lowercased and stopword-filtered) for ``Zeroshot'' (Sec.~\ref{section_exp1}) and ``Fewshot'' (Sec.~\ref{section_label_analysis}) relabeling. There were 11051 relabeled trajectories.}
\label{appdx-exp1-fs-zs-unigram-statistics}
\end{center}
\end{figure}

\begin{table}
\centering
\begin{tabular}{lr}
\toprule
Zeroshot Caption &  Frequency \\
\midrule
a basketball                               &            875 \\
a banana                                   &            525 \\
it is a basketball                         &            480 \\
a cube                                     &            364 \\
a pear                                     &            319 \\
a chair                                    &            317 \\
it is a banana                             &            303 \\
it is a chair                              &            242 \\
a car                                      &            221 \\
it is a cube that is a child of the camera &            209 \\
a plane                                    &            202 \\
it is a cube that is a child of a camera   &            198 \\
a box                                      &            193 \\
it is a car                                &            172 \\
a table                                    &            171 \\
a racket                                   &            154 \\
its a basketball                           &            152 \\
it is a table                              &            139 \\
it is a pear                               &            122 \\
it is a basketball hoop                    &            119 \\
a carrot                                   &            108 \\
it is a plane                              &             91 \\
it is a cube                               &             88 \\
a tennis racket                            &             83 \\
its a banana                               &             81 \\
\bottomrule
\end{tabular}
\caption{Top 25 Zeroshot captions (lowercased and punctuation-stripped). There were 11051 total relabeled trajectories (Section~\ref{section_exp1}).}
\label{appdx-exp1-zeroshot-labels}
\end{table}

\begin{table}
\centering
\begin{tabular}{lr}
\toprule
Zeroshot Caption &  Frequency \\ \\
\midrule
its a toy                                          &                 7704 \\
it is a toy                                        &                 1022 \\
its both                                           &                  676 \\
both                                               &                  400 \\
its a food toy                                     &                   63 \\
its a toy but its also food                        &                   30 \\
its food                                           &                   22 \\
it is both                                         &                   20 \\
it is a toy but it is also food                    &                    9 \\
its a toy but its also a food                      &                    8 \\
its a food                                         &                    7 \\
its a toy but it can be used as a food             &                    5 \\
its a toy but it is also food                      &                    3 \\
food                                               &                    3 \\
its a toy but its not a toy                        &                    2 \\
its food but its also a toy                        &                    2 \\
its a toy carrot                                   &                    2 \\
its food but its not edible                        &                    2 \\
its a toy but its a toy that you can eat           &                    2 \\
its a toy but it can be used as food               &                    2 \\
its a food simulator                               &                    2 \\
its a toy its a food toy                           &                    1 \\
its a toy but it can be used as food if you wan... &                    1 \\
its a toy that is also food                        &                    1 \\
its a food that is also a toy                      &                    1 \\
its a toy that looks like food                     &                    1 \\
its a toy its a toy                                &                    1 \\
it is a food                                       &                    1 \\
its food for your mind                             &                    1 \\
its a toy but it can be used to feed your pet      &                    1 \\
its a toy but its not a toy you can play with      &                    1 \\
its a food but its not edible                      &                    1 \\
it is a food toy                                   &                    1 \\
its a toy but it can be a food too                 &                    1 \\
it is a toy but the food is coming soon            &                    1 \\
it is a toy it is not edible                       &                    1 \\
its a toy but its not a ball                       &                    1 \\
\bottomrule
\end{tabular}
\caption{All ``food or toy'' ``Zeroshot'' captions (lowercased and punctuation-stripped). There were 10002 total relabeled trajectories (Section~\ref{section_exp3}).}
\label{appdx-foodtoy-zeroshot-labels}
\end{table}

\begin{table}
\centering
\begin{tabular}{lr}
\toprule
Fewshot-1 Caption &  Frequency \\ \\
\midrule
toy  &                 6186 \\
food &                 3816 \\
\bottomrule
\end{tabular}
\caption{All ``food or toy'' ``Fewshot-1'' captions (lowercased and punctuation-stripped). There were 10002 total relabeled trajectories (Section~\ref{section_exp3}).}
\label{appdx-foodtoy-fewshot-labels}
\end{table}

\begin{table}[h]
\centering
\begin{tabular}{rlllrrrrr}
  \hline
 & Effect & Group & Term & Estimate & Std. Error & Statistic & DOF & P Value \\ 
  \hline
1 & fixed &  & (Intercept) & -0.70 & 0.11 & -6.43 & 36.50 & 1.74e-07 *** \\ 
  2 & fixed &  & Label Precision & 1.44 & 0.13 & 11.50 & 35.76 & 1.43e-13 *** \\ 
  3 & fixed &  & Label Accuracy & 0.16 & 0.05 & 3.42 & 35.88 & 0.00159 ** \\ 
  4 & ran\_pars & Task & sd\_\_(Intercept) & 0.05 &  &  &  &  \\ 
  5 & ran\_pars & Residual & sd\_\_Observation & 0.05 &  &  &  &  \\ 
   \hline
\end{tabular}
\caption{Regressing task performance on label accuracy and precision shows that the effects of label precision far outweigh those of accuracy (Section~\ref{section_label_analysis}).}
\label{appdx-regression-results}
\end{table}

\FloatBarrier

\section{OWL-ViT Comparisons}
\label{appdx-sec-owlvit}

We compared OWL-ViT~\cite{minderer2022owlvit} based relabeling to Flamingo in our Experiment 1 (Section~\ref{section_exp1}). We provided OWL-ViT with the list of 10 objects used in the experiment (plane, basketball, chair, table…) and took the highest-confidence detection. We found that OWL-ViT performed poorly: on a subset of 300 trajectories, it predicted the held object only 7.2\% of the time, compared to Flamingo’s zero-shot 54.8\% performance. OWL-ViT only ever predicted a limited subset of objects and achieved extremely low precision on these (Table~\ref{tab:appdx-owl-vit-summary}). Flamingo had higher accuracy than OWL-ViT across 9/10 objects, often by wide margins (Table~\ref{tab:appdx-owl-vit-accuracy}).

\begin{table}[h]
\centering
\begin{tabular}{lrr}
\toprule

OWL-ViT Label & Count &  Precision \\
\midrule
basketball     &     39 &         0.10 \\
book           &     69 &         0.12 \\
chair          &     44 &         0.07 \\
table          &    140 &         0.04 \\
\bottomrule
\end{tabular}
\caption{OWL-ViT consistently predicted a small subset of the 10 object labels, achieving very low precision.}
\label{tab:appdx-owl-vit-summary}
\end{table}

\begin{table}[h]
\centering
\begin{tabular}{lrrr}
\toprule
Object &  Count &  OWL-ViT Accuracy &  Flamingo Accuracy \\
\midrule
banana            &     34 &         0.00 &              0.76 \\
basketball        &     47 &         0.09 &              1.00 \\
book              &     27 &         0.30 &              0.11 \\
car               &     23 &         0.00 &              0.65 \\
carrot            &     28 &         0.00 &              0.14 \\
chair             &     24 &         0.12 &              0.58 \\
pear              &     14 &         0.00 &              0.64 \\
plane             &     43 &         0.00 &              0.58 \\
racket            &     34 &         0.00 &              0.24 \\
table             &     18 &         0.33 &              0.50 \\
\bottomrule
\end{tabular}
\caption{Flamingo achieved substantially higher accuracy than OWL-ViT on 9/10 objects.}
\label{tab:appdx-owl-vit-accuracy}
\end{table}

Our OWL-ViT implementation performed as expected on real-world images, so we hypotheize its poor performance here is due to domain shift from real-world to Unity-generated environments. Notably, Flamingo often produced captions which reflected this domain shift (e.g. ``a 3d model of a car'', Fig.~\ref{fig-exp1-summary}). It may thus be possible to improve OWL-ViT’s performance by changing the target labels to reflect the objects’ idiosyncratic visual appearances. However, this necessity reflects the brittleness of the object-detection approach, compared to zero-shot generative captioning.

\section{Testing Preference Generalization}
\label{appdx-sec-recoloring-generalization}

We investigated whether the system is able to generalize to other instances of preferred objects (Section~\ref{section_exp4}). Specifically, we kept the few-shot prompt the same, but varied the color of the objects in the environment by permuting the toys’ colors. We then examined the VLM’s ability to re-label them. We found that across all 5 toys, for both ``aligned'' and ``arbitrary'' preference structures, the VLM readily generalized to the new colors (Table~\ref{tab:appdx-recoloring-generalization-summary}). The VLM’s accuracy is consistent across objects (Table~\ref{tab:appdx-recoloring-generalization-individual}).

\begin{table}[h]
\centering
\begin{tabular}{llr}
\toprule
Preference Structure & Coloring &  Accuracy \\
\midrule
   Aligned &  Canonical &     0.91 \\
    Aligned &  Recolored &     0.92 \\
  Arbitrary &  Canonical &     0.71 \\
  Arbitrary &  Recolored &     0.70 \\
\bottomrule
\end{tabular}
\caption{The VLM readily generalized to relabeling recolored objects with ad-hoc category structures.}
\label{tab:appdx-recoloring-generalization-summary}
\end{table}

\begin{table}[h]
\centering
\begin{tabular}{lllr}
\toprule
preferences & unity\_object\_name &                  color &  correct \\
\midrule
    Aligned &               Car & Canonical - aquamarine &     0.91 \\
    Aligned &               Car &        Recolored - red &     0.92 \\
    Aligned &              Dice &      Canonical - white &     0.97 \\
    Aligned &              Dice &     Recolored - purple &     0.98 \\
    Aligned &             Plane &     Canonical - orange &     0.85 \\
    Aligned &             Plane & Recolored - aquamarine &     0.87 \\
    Aligned &             Robot &     Canonical - purple &     0.96 \\
    Aligned &             Robot &     Recolored - orange &     0.92 \\
    Aligned &             Train &       Canonical - none &     0.87 \\
    Aligned &             Train &      Recolored - white &     0.92 \\
  Arbitrary &               Car & Canonical - aquamarine &     0.46 \\
  Arbitrary &               Car &        Recolored - red &     0.47 \\
  Arbitrary &              Dice &      Canonical - white &     0.94 \\
  Arbitrary &              Dice &     Recolored - purple &     0.96 \\
  Arbitrary &             Plane &     Canonical - orange &     0.85 \\
  Arbitrary &             Plane & Recolored - aquamarine &     0.80 \\
  Arbitrary &             Robot &     Canonical - purple &     0.70 \\
  Arbitrary &             Robot &     Recolored - orange &     0.64 \\
  Arbitrary &             Train &       Canonical - none &     0.59 \\
  Arbitrary &             Train &      Recolored - white &     0.55 \\
\bottomrule
\end{tabular}
\caption{The VLM readily generalized across all objects, achieving comparable accuracy on original and recolored versions.}
\label{tab:appdx-recoloring-generalization-individual}
\end{table}


\end{document}